\pdfoutput=1

\documentclass[11pt]{article}

\usepackage[preprint]{acl}

\usepackage{times}
\usepackage{latexsym}

\newcommand{\nameshort}{FastMem}

\usepackage[T1]{fontenc}

\usepackage[utf8]{inputenc}

\usepackage{microtype}

\usepackage{inconsolata}

\usepackage{graphicx}
\usepackage{amsmath, bm}
\usepackage{cleveref}
\usepackage{amssymb}
\usepackage{booktabs}
\usepackage{hhline}
\usepackage{todonotes}
\usepackage{tikz}
\crefname{figure}{Fig.}{Figs.}
\crefname{table}{Tab.}{Tabs.}
\crefname{equation}{Eq.}{Eqs.}
\crefname{section}{Sec.}{Secs.}
\crefname{appendix}{Sec.}{Secs.}
\crefname{subsection}{Sec.}{Secs.}
\crefname{subsubsection}{Sec.}{Secs.}
\newcommand{\bftab}{\fontseries{b}\selectfont}
\usepackage{multirow}
\usepackage{enumitem}
\usepackage{fancyvrb}
\usepackage{geometry}
\usepackage[most]{tcolorbox}
\usepackage{listings}
\title{\includegraphics[height=1.5em]{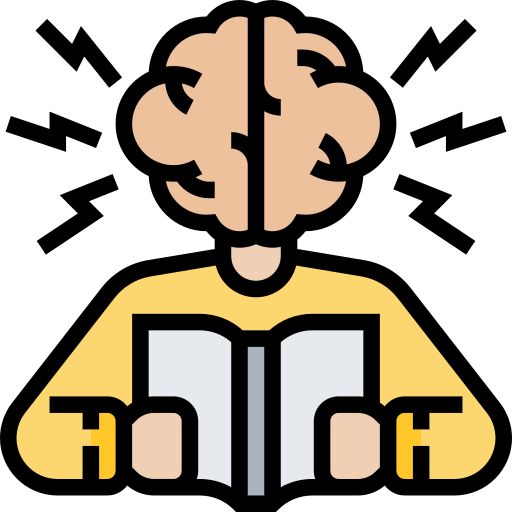} \hspace{0.1em}FastMem: \underline{Fast} \underline{Mem}orization of Prompt Improves Context Awareness of Large Language Models}

\author{
 \textbf{Junyi Zhu\thanks{Equal Contribution}\textsuperscript{1}},
 \textbf{Shuochen Liu$^*$\textsuperscript{2}},
 \textbf{Yu Yu\textsuperscript{3}},
 \textbf{Bo Tang\thanks{\textbf{Correspondence:} \href{mailto:tangb@iaar.ac.cn}{tangb@iaar.ac.cn}}\textsuperscript{2,3}},
 \textbf{Yibo Yan\textsuperscript{4}},
 \textbf{Zhiyu Li\textsuperscript{3}},
\\
 \textbf{Feiyu Xiong\textsuperscript{3}},
 \textbf{Tong Xu\textsuperscript{2}},
 \textbf{Matthew B. Blaschko\textsuperscript{1}}
\\
\\
 \textsuperscript{1}ESAT-PSI, KU Leuven,
 \\
 \textsuperscript{2}	University of Science and Technology of China,
\\
 \textsuperscript{3}	Institute for Advanced Algorithms Research, Shanghai,
 \\
 \textsuperscript{4} National University of Singapore
}

\begin{document}
\maketitle
\begin{abstract}
Large language models (LLMs) excel in generating coherent text, but they often struggle with context awareness, leading to inaccuracies in tasks requiring faithful adherence to provided information. We introduce \textbf{FastMem}, a novel method designed to enhance \textit{instruction fine-tuned LLMs}' context awareness through fast memorization of the prompt. \textbf{FastMem} maximizes the likelihood of the prompt before inference by updating only the last Feed-Forward Network (FFN) module. This targeted approach ensures efficient optimization without overfitting, significantly improving the model's ability to comprehend and accurately follow the context. Our experiments demonstrate substantial gains in reading comprehension, text summarization and adherence to output structures. For instance, \textbf{FastMem} improves the accuracy of Llama 3-8B-Inst on the NQ-SWAP dataset from 59.1\% to \textbf{71.6\%}, and reduces the output structure failure rate of Qwen 1.5-4B-Chat from 34.9\% to \textbf{25.5\%}. Extensive experimental results highlight \textbf{FastMem}'s potential to offer a robust solution to enhance the reliability and accuracy of LLMs in various applications. Our code is available at: \url{https://github.com/IAAR-Shanghai/FastMem}.
\end{abstract}

\section{Introduction}
\label{sec:intro}
Large language models (LLMs) function within a versatile input space defined by prompts. These prompts set the task context, provide background information for answering questions, and offer examples for in-context learning~\citep{brown2020language,wei2022finetuned,t5,sanh2022multitask,radford2019language}.

Despite their remarkable utility, \textit{LLMs have difficulty in fully understanding and following the provided context, which we refer as the ability of context awareness}. In this paper, we consider both the instruction in the prompt, such as query or predefined output format, and the reference provided by the user or retrieval system as the context. A major issue with context awareness is maintaining faithfulness to the contextual information. For instance, in summarization tasks, LLMs may fabricate entities or relationships that are not present in the original reference~\citep{pagnoni-etal-2021-understanding,cao-etal-2020-factual,maynez-etal-2020-faithfulness}. This issue also reduces the effectiveness of Retrieval-Augmented Generation (RAG)~\citep{liu2023evaluating,Chen_Lin_Han_Sun_2024}, as the LLMs may falsify the information in references found by RAG when answering the question. Another aspect is adhering to a predefined output structure, which is necessary for many applications ~\citep{liu2024we}.

\begin{figure}[t]
    \centering
    \includegraphics[width=\linewidth,scale=1.00]{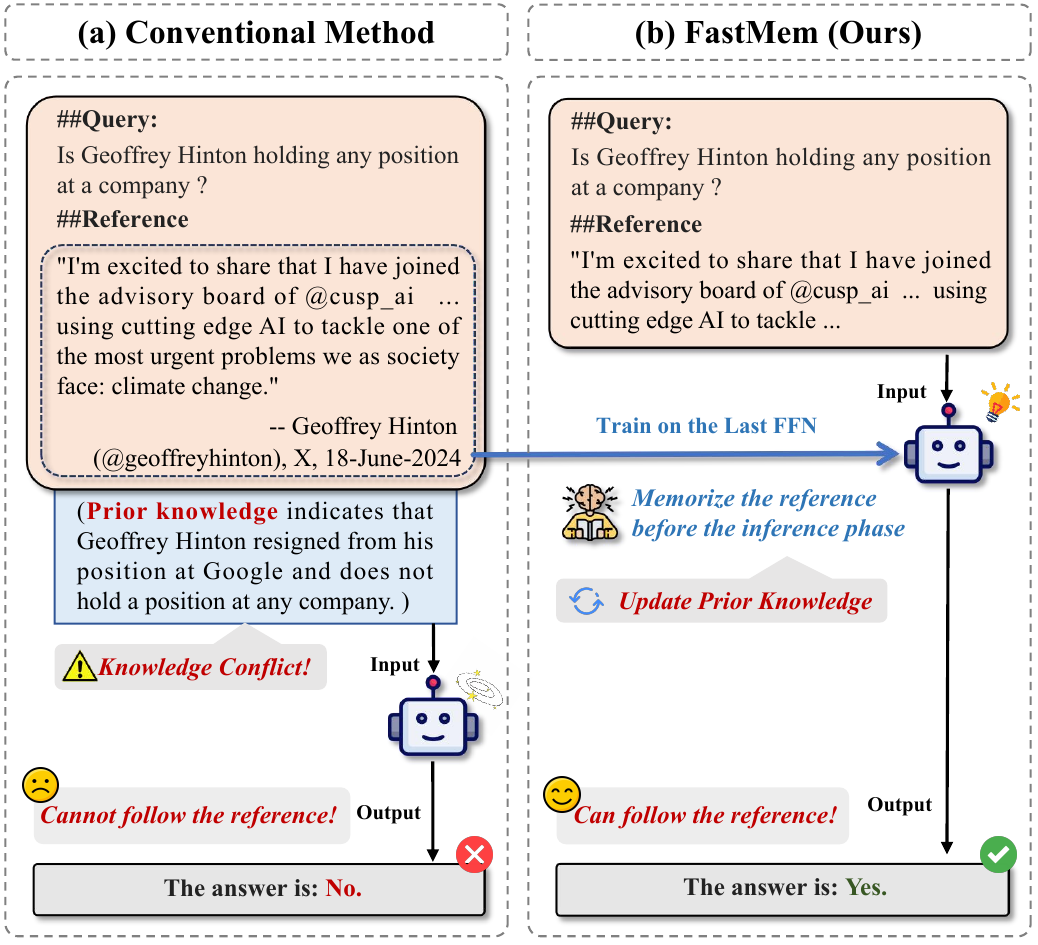}
    \caption{An illustrative example where an LLM cannot follow the reference due to its conflict with the LLM's prior knowledge (a). In such cases, LLM often exhibits high perplexity on the conflicted information. Our proposed FastMem addresses this issue by enabling the model to memorize the reference text (thereby reducing perplexity) before conducting inference (b).}
    \label{fig:comparison}
\end{figure}

LLMs undergo pre-training and fine-tuning. The pre-training stage involves learning on a large corpus of text to parameterize extensive knowledge into the network~\citep{Radford2018ImprovingLU,radford2019language}.
After pre-training, various abilities of the model are further elicited by fine-tuning on more specific tasks or via reinforcement learning with human feedback ~\citep{mishra-etal-2022-cross,ouyang2022training}. Although fine-tuning improves LLMs context awareness, previous work has shown that LLMs may not always faithfully follow the information provided in the context when answering questions~\citep{longpre-etal-2021-entity,zhou2023contextfaithful,wu2024faithful, liu2023evaluating}. This issue becomes particularly pronounced when there is a conflict between the ingrained parametric knowledge and the contextual information provided during inference. For instance, such conflicts can happen when the pre-training dataset is outdated or lacks recent information. Additionally, the diversity of pre-training corpus can result in a mass-seeking effect, where the model assigns non-zero probabilities to many tokens~\citep{hallu_survey}. Mass-seeking behavior of LLMs can exacerbate the challenge of context awareness.

These findings imply that LLMs perform worse when they have high perplexity regarding the context. We also verify that models perform worse in extractive question answering when the perplexity of the answer span in the context is high (see \cref{sec:motivation}). Inspired by these insights, we propose a novel approach named FastMem to improve context awareness via fast memorization of prompts, as depicted in \cref{fig:comparison}. Specifically, we adapt an \textit{instruction fine-tuned model} by maximizing the prompt likelihood before generation. To ensure the efficiency of the optimization and prevent overfitting to small text snippets, we update only the last FFN module, achieving significant improvements with minimal computational cost.
Our contributions are summarized as follows:

\vspace{+0.4em}
\noindent\textbf{1}) We propose a novel approach, \nameshort{}, that integrates memorization into the inference process. Specifically, \nameshort{} employs a pretrain-like objective for an instruction fine-tuned model, optimizing it to memorizing the context provided in the prompt (i.e., reduce perplexity on the context).

\vspace{+0.4em}
\noindent \textbf{2}) To avoid resource-intensive full fine-tuning, we explore optimizing a subset of parameters and find that tuning the last FFN module yields significant improvements and can be done quickly. \nameshort{} can be completed within a few seconds and without increase in peak memory usage.

\vspace{+0.4em}
\noindent\textbf{3}) \nameshort{} can significantly enhance the model's reading comprehension abilities when conflicts between parametric knowledge and contextual information are present. For instance, \nameshort{} improves Llama 3-8B-Instruct's accuracy on NQ-SWAP~\citep{longpre-etal-2021-entity} from 59.1\% to \textbf{71.6\%}.

\vspace{+0.4em}
\noindent\textbf{4}) \nameshort{} can also be applied to emphasize the output format. For instance, we find that Qwen 1.5-4B-Chat occasionally fails to follow the predefined output structure. However, when \nameshort{} is used to memorize the instruction for the output format first, the failure rate of the output structure is reduced from 34.9\% to \textbf{25.5\%}, and overall performance is significantly improved.

\section{Related Work}

\begin{figure*}[t]
    \centering
    \includegraphics[width=\linewidth]{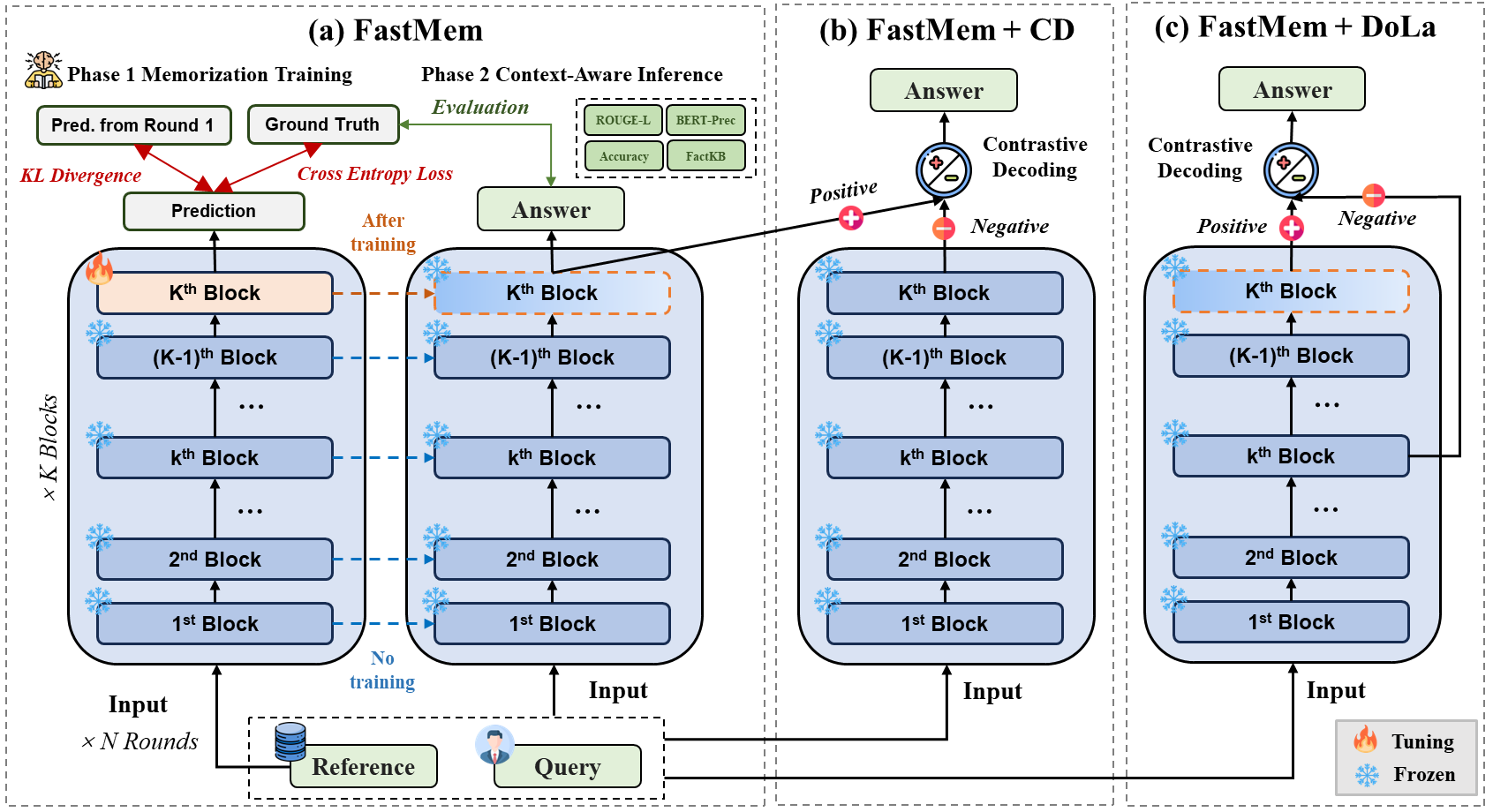}
    \caption{Overall frameworks of our proposed FastMem and its integration with decoding strategies.}
    \label{fig:main}
\end{figure*}

\noindent\textbf{Fine Tuning.} 
Instruction tuning typically enhances a model's ability to understand and respond to contextual information~\citep{brown2020language,ouyang2022training,mishra-etal-2022-cross,sanh2021multitask}. Many studies have explored tuning language models on dedicated datasets to further enhance their context awareness and improve aspects such as faithfulness and factual accuracy~\citep{goyal-durrett-2021-annotating,cao-wang-2021-cliff,wan-bansal-2022-factpegasus}. Unlike these methods, \nameshort{} optimizes on individual data examples and does not require curating a specific dataset. Additionally, \textit{\nameshort{} is applied upon instruction fine-tuned models to further enhance their context awareness.}

\vspace{+0.3em}
\noindent\textbf{Decoding Strategy.}
Many decoding strategies have been proposed to improve context awareness. One mainstream approach leverages the evaluative capabilities of LLMs to enhance the quality and factuality of their generated text~\citep{madaan2023selfrefine,2024selfcontradictory,saunders2022self}. Another approach applies the concept of contrast to strengthen context awareness~\citep{li-etal-2023-contrastive,zhang2023alleviating,chuang2024dola,shi2023trusting}.\textit{ Ours \nameshort{} and these decoding strategies are both inference-time methods, while \nameshort{} is compatible with existing decoding strategies}. In this work, we also investigate incorporating two decoding strategies to further enhance the effectiveness of our method.

\vspace{+0.3em}
\noindent\textbf{Knowledge Editing.} 
Knowledge editing methods update and correct factual information within language models by modifying the model's parametric knowledge~\citep{de-cao-etal-2021-editing,NEURIPS2022_6f1d43d5,dai-etal-2022-knowledge,huang2023transformerpatcher}. Both knowledge editing and \nameshort{} involve training on individual data examples, but knowledge editing requires input-output pairs (e.g., question and answer), while \textit{\nameshort{} uses only the input, making it more widely applicable. Additionally, knowledge editing is not as efficient as \nameshort{} to be applied at inference time}.

\vspace{+0.3em}
\noindent\textbf{Test-Time Training.} 
Test-Time Training (TTT) was initially proposed for vision tasks to adapt models for out-of-distribution data using psuedo-labels~\citep{pmlr-v119-sun20b,pmlr-v151-bartler22a}. In contrast, \textit{\nameshort{} can improve model performance when the train and test data distributions are inconsistent or consistent}. Our experiments reveal that a negative correlation between the performance and perplexity with respect to the context exits in LLMs, as discussed in \cref{sec:motivation}. This negative correlation potentially leads to a more general applicability of \nameshort{}. Recently, \citet{hardt2024testtime} studied TTT for language models. However, they only consider pretrained models, in which raw text for TTT can be directly fed to the network. If instruction fine-tuned models follow the same method without modification, they risk being corrupted based on our experiments. \textit{In \nameshort{} we address this problem with dedicated control tokens. Furthermore, we explore the reduction of latency and achieve real-time application of our method}. Thus, \nameshort{} distinguishes itself from existing TTT methods.

\section{Method}
In this section, we elaborate on our method, \nameshort{}. In \cref{sec:notation}, we clarify the notations. In \cref{sec:motivation}, we present the motivation behind our method. In \cref{sec:fastmem}, we explain how we develop our method to be robust and efficient. In \cref{sec:combine}, we explore combining \nameshort{} with existing decoding strategies. Finally, in \cref{sec:usecase}, we demonstrate the use cases. An illustration of our approach and its variants is provided in \cref{fig:main}.

\subsection{Notations}
\label{sec:notation}
Let $\bm x$ denote the memorization text consisting of a sequence of tokens. Let $\bm x_{t}$ denote the $t$-th token in the sequence $\bm x$. The language model with $K$ transformer blocks is parameterized by $\bm \theta^k,\, k=1\ldots K$, and specifically $\bm \theta^K_{FFN}$ denotes the parameters of the last FFN module. The output of the model is given by the conditional probability distribution $P_{\bm \theta}(\bm x_{t} | \bm x_{<t})$, where $\bm x_{<t}$ denotes the sequence of tokens preceding $\bm x_{t}$.

\begin{figure}[t]
    \centering
    \includegraphics[width=0.8\columnwidth]{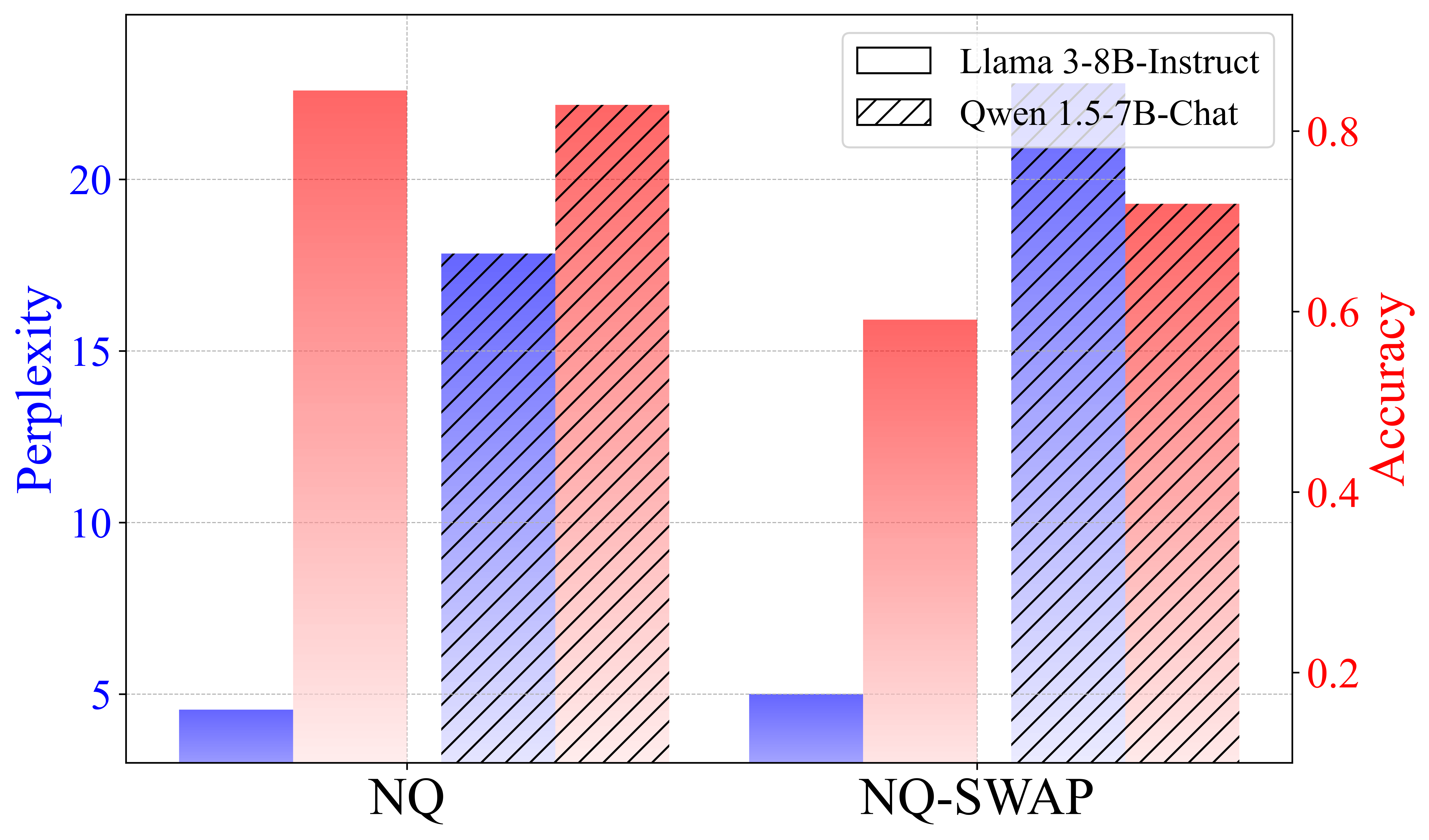}
    \vspace{-.5em}
    \caption{Model performance and perplexity on NQ and NQ-SWAP datasets.}
    \label{fig:motivation}
    \vspace{-.5em}
\end{figure}

\subsection{Motivation}
\label{sec:motivation}
As introduced in \cref{sec:intro}, conflicts between parametric knowledge and contextual information, as well as mass-seeking behavior are supposed to hamper the context awareness. To verify that, we compare the performance of Llama 3-8B-Instruct~\citep{meta2023llama3} and Qwen 1.5-7B-Chat~\citep{bai2023qwen} on two datasets: Natural Questions (NQ)~\citep{kwiatkowski-etal-2019-natural} and NQ-SWAP. NQ is constructed based on Wikipedia, which is included in many pre-training corpora. Therefore, the information from NQ has likely been parameterized into Llama 3 and Qwen 1.5. NQ-SWAP is constructed from NQ using the method of \citet{longpre-etal-2021-entity}, where we swap the answer strings, which are entities, with random similar entities from the corpus (cf.\ \cref{sec:app:nqswap}). This process creates new context-answer pairs that the model has not observed before.

We measure the model's perplexity regarding the minimal span of tokens in the context containing the answer string. As shown in \cref{fig:motivation}, we observe that models exhibit higher perplexity on NQ-SWAP compared to NQ, proving that the information in NQ belongs to the model's parametric knowledge. Additionally, the model suffers an obvious utility drop on NQ-SWAP, although the differences between the two datasets are merely short strings involving entities, indicating a negative correlation between perplexity and performance.

The negative correlation suggests that optimizing the model to memorize the context (thereby reducing the perplexity) before inference could be beneficial. This approach can potentially address the commonly hypothesized causes of context-awareness issues, as introduced in~\cref{sec:intro}. First, reducing the perplexity on context can alleviate potential conflicts between parametric knowledge and contextual information. Second, by increasing the model's certainty regarding the text in the context, the mass-seeking effect will be reduced.

\subsection{Efficient and robust fast memorization of prompt}
\label{sec:fastmem}
Since the memorization approach is applied during inference time, it should not significantly increase computation complexity. Additionally, memorizing a single example may degrade the model's general language ability. Addressing these issues are crucial for applicability.
Next, we elaborate on how we develop \nameshort{} to be robust and efficient.

\subsubsection{Memorization objective}
\label{sec:obj}
To memorize an individual example $\bm x$, we adopt the next token prediction objective, which is defined as minimizing the negative log-likelihood of the true next token $\bm x_{t}$ given the previous tokens in the sequence. This can be formulated as:
\begin{equation}
\label{eq:ntp}
\mathcal{L}_{\text{NTP}}(\bm x; \bm \theta) = - \sum_{t=1}^{T} \log P_{\bm \theta}(\bm x_{t} |\bm x_{<t}),
\end{equation}
where $T$ is the length of the sequence $\bm x$.

\paragraph{KL Divergence Regularization.}

To avoid overfitting on the memorized text and degradation of language ability, we regularize the deviation from the model's initial output using a Kullback-Leibler (KL) divergence term. Let $P_{\bm \theta_0}(\bm x_{t} | \bm x_{<t})$ denote the initial output distribution of the model before optimization, and $P_{\bm \theta_N}(\bm x_{t} | \bm x_{<t})$ denote the output distribution after $N$ training iterations. The KL divergence regularization term is defined as:
\begin{align}
\nonumber
&\mathcal{L}_{\text{KL}}(\bm x; \bm \theta_N) \\
&= \sum_{t=1}^{T} \mathrm{KL} \left( P_{\bm \theta_0}(\bm x_{t} | \bm x_{<t}) \parallel P_{\bm \theta_N}(\bm x_{t} | \bm x_{<t}) \right).
\end{align}

\paragraph{Last-FFN Adaptation.} Our experiments show that optimizing only the last FFN module $\bm \theta_{FFN}^K$ yields the best improvement (see \cref{sec:ablation}). Previous studies have observed that the FFN modules store knowledge~ \citep{geva-etal-2021-transformer,dai-etal-2022-knowledge}. Moreover, many works in knowledge editing focus on modifying the last few FFN modules ~\citep{huang2023transformerpatcher,mitchell2022fast}. Additionally, another major benefit of last-FFN adaptation is efficiency. By saving the hidden state outputs from the attention layer of the last transformer block and reusing them to optimize the last FFN module, the optimization is essentially performed on a small MLP network, making the process quick and memory-efficient. Therefore, our \nameshort{} method fine-tunes only the last FFN module.

\paragraph{Overall Objective.}
The overall objective for optimizing the model on an individual example combines the next token prediction objective and the KL divergence regularization. Additionally, only the last FFN module is updated while other parameters are frozen:
\begin{equation}
\nonumber
\bm \theta_{FFN}^K \rightarrow \
\underset{\bm \theta_{FFN}^K}{arg\,min}\;\mathcal{L}_{\text{NTP}}(\bm x; \bm \theta) + \lambda \mathcal{L}_{\text{KL}}(\bm x; \bm \theta),
\end{equation}
where $\lambda$ is a hyperparameter that controls the trade-off between the next token prediction loss and the KL divergence regularization.

\subsubsection{Memorization control tokens}
The memorization objective defined in \cref{sec:obj} is similar to the pre-training objective. However, straightforwardly optimizing an instruction fine-tuned model on the text as in pre-training causes the model to degrade and produce chaotic responses. We hypothesize that this optimization disrupts the model's understanding of the chat template. Therefore, it is crucial to apply the chat template during memorization, and the best approach we found is to use \textit{control tokens for the machine response}, with examples provided in \cref{fig:control_tokens,fig:control_tokens_qwen}. Intuitively, the control tokens in the memorization template guide the model to repeat certain text in its role. When computing the next token prediction loss in \cref{eq:ntp}, we will mask out the control tokens, except for the end-of-text token, e.g., \textcolor{blue}{\textless\textbar eot\textbar\textgreater}. Including an end-of-text token in memorization can prevent the model from generating endlessly afterward.

\subsection{Integrating decoding strategies with \nameshort{}}
\label{sec:combine}
\nameshort{} as described in \cref{sec:fastmem} can already improve context awareness. However, given its potential to strengthen the model, we further explore whether it can be integrated with existing decoding strategies to achieve higher performance. Specifically, we investigate the integration of two contrastive decoding strategies.

\paragraph{\nameshort{} + CD.}
\citet{li-etal-2023-contrastive} propose contrastive decoding (CD) which considers the difference between the likelihoods predicted by a strong model and a weak model. Following the concept of CD, we make the prediction of \nameshort{} contrasted with the original prediction made by the model before any memorization took place, as depicted in \cref{fig:main}(b). This is done by subtracting the original prediction probabilities \( P_{\bm \theta_0}(\bm x_{t} | \bm x_{<t}) \) from the current probabilities \( P_{\bm \theta_N}(\bm x_{t} | \bm x_{<t}) \). This subtraction aims to emphasize the differences introduced by the memorization process and mitigate any overfitting by highlighting where the model’s behavior has significantly changed. The modified prediction $\tilde{P}_{\bm \theta_N}(\bm x_{t} | \bm x_{<t})$ can be expressed as:
\begin{align}
    \nonumber
    softmax(&(1+\alpha)\log P_{\bm \theta_N}(\bm x_{t} | \bm x_{<t}) -\\
    &\alpha \log P_{\bm \theta_0}(\bm x_{t} | \bm x_{<t})), 
\end{align}
where $\alpha$ is introduced to adjust the contrastive effect following \citet{shi2023trusting}.

\paragraph{\nameshort{} + DoLa.}
In the second strategy, the final prediction is contrasted with an intermediate representation within the network, following the DoLa framework \citep{chuang2024dola}. Specifically, we take the output from an earlier transformer block \( k \) where \( k < K \), and pass it through the final projection layer and softmax function to obtain an early-exit prediction \( P_{\bm{\theta_0}^{\{1:k\}}}(\bm{x}_t | \bm{x}_{<t}) \), as depicted in \cref{fig:main}(c). The final output \( P_{\bm{\theta_N}}(\bm{x}_t | \bm{x}_{<t}) \) is then adjusted by subtracting the early-exit prediction. This approach leverages the hierarchical nature of transformer models, where different layers capture different levels of abstraction. By contrasting the final output with an early-exit prediction, we can emphasize the knowledge learned and output by the last fine-tuned FFN. The adjusted prediction can be formalized as:
\begin{align}
    \nonumber
    &\tilde{P}_{\bm\theta_N}(\bm{x}_t | \bm{x}_{<t}) = \text{softmax}(\mathcal{F}(P_{\bm\theta_N}, P_{\bm\theta_N^{\{1:k\}}})), \\
    \nonumber
    &\mathcal{F}(\cdot) = \begin{cases} 
        \log \left( \frac{P_{\bm\theta_N}(\bm{x}_t | \bm{x}_{<t})}{P_{\bm\theta_N^{\{1:k\}}}(\bm{x}_t | \bm{x}_{<t})} \right) & \text{if } \bm{x}_t \in \mathcal{V}_{\text{head}} \\
        -\infty & \text{otherwise}
    \end{cases},
\end{align}
where \(\mathcal{V}_{\text{head}}\) is a set of tokens having high enough output probabilities in \( P_{\bm\theta_N}(\bm{x}_t | \bm{x}_{<t}) \). The \( k \)-th intermediate block is selected as follows:
\begin{align}
    \nonumber
    &k = arg\,max_{k \in M} d(P_{\bm\theta_N}, P_{\bm\theta_N^{\{1:k\}}}), \\
    \nonumber
    &d(\cdot) = JSD(P_{\bm\theta_N}(\bm{x}_t | \bm{x}_{<t}) \| P_{\bm\theta_N^{\{1:k\}}}(\bm{x}_t | \bm{x}_{<t})),
\end{align}
where JSD represents Jensen–Shannon divergence and $M$ is a set of candidate blocks. These equations identify the early-exit prediction with the maximal difference from the final prediction in the predefined set of the intermediate blocks.

\begin{figure}[t]
    \centering
\begin{tcolorbox}[colback=white,colframe=black, enhanced jigsaw, listing only, listing options={basicstyle=\rmfamily}]

\textbf{\textit{Memorization}}: \textcolor{blue}{\textless\textbar start\_header\_id\textbar\textgreater assistant\textless\textbar end\_header\_id\textbar\textgreater}

\textcolor{gray}{\texttt{<reference>}}
\textcolor{blue}{\textless\textbar eot\_id\textbar\textgreater}

\textbf{\textit{Inference}}: 

\textcolor{blue}{\textless\textbar begin\_of\_text\textbar\textgreater\textless\textbar start\_header\_id\textbar\textgreater user\textless\textbar}\\
\textcolor{blue}{end\_header\_id\textbar\textgreater} \\ 
        Please extract from the reference the span that best answers the question and provide the answer in the following format: "The answer is: ...".\\
        Reference: \textcolor{gray}{\texttt{<reference>}} \\
        Question: \textcolor{gray}{\texttt{<question>}} \textcolor{blue}{\textless\textbar eot\_id\textbar\textgreater} \\
        \textcolor{blue}{\textless\textbar start\_header\_id\textbar\textgreater assistant\textless\textbar end\_header\_id\textbar\textgreater}
\end{tcolorbox}
\vspace{-0.5em}
\caption{Q\&A template for Llama 3. Control tokens are highlighted in blue. Keys are colored in gray and are replaced by corresponding content from the dataset.} 
\label{fig:control_tokens}
\end{figure}

\begin{figure}[t]
    \centering
\begin{tcolorbox}[colback=white,colframe=black, enhanced jigsaw, listing only, listing options={basicstyle=\rmfamily}]

\textbf{\textit{Memorization}}: \textcolor{blue}{\textless\textbar im\_start\textbar\textgreater assistant}\\
Please extract from the reference the span that best answers the question and provide the answer in the following format: "The answer is: ...".\\
Reference: \textcolor{gray}{\texttt{<reference>}}
\textcolor{blue}{\textless\textbar im\_end\textbar\textgreater}
\end{tcolorbox}
\vspace{-0.5em}
\caption{Q\&A memorization template for Qwen 1.5-4B-Chat. Instructions are embedded within the memorization text to improve the output structure. For inference template, refer to \cref{fig:control_tokens}. Control tokens are highlighted in blue, while keys are shown in gray and are replaced with corresponding content from the dataset.}
\label{fig:control_tokens_qwen}
\end{figure}
\subsection{Use cases}
\label{sec:usecase}
By incorporating different segments of the prompt into memorization, we can realize many utilities of \nameshort{}. In this work, we present two significant use cases where \nameshort{} demonstrates its value.

\paragraph{Improving Context Understanding and Faithfulness.}
When using RAG for question answering or summarizing text given by the user, \nameshort{} can be applied to make the model memorize the reference or the given text first (see \cref{fig:control_tokens}). We find that \nameshort{} enhances the model's understanding of the text and improves the accuracy of its responses. In scenarios where conflicts between parametric knowledge and contextual information are present, \nameshort{} significantly improves context faithfulness, thereby enhancing the performance.

\paragraph{Enhancing Instruction Following.} If instructions in the prompt are incorporated into memorization (see \cref{fig:control_tokens_qwen}), \nameshort{} can also help the model adhere to the given instructions. For instance, we find that \nameshort{} aids in making the model output follow the structure defined in the prompt.

\section{Experiments}
In this section, we evaluate FastMem on Q\&A and summarization tasks that require LLMs to respond based on the given context.

\subsection{Experimental configurations
}
\paragraph{Baseline.} 
Our experiments include evaluations of popular open-source instruction fine-tuned LLMs from the Llama 3 series (8B and 70B) and the Qwen 1.5 series (4B, 7B, and 14B). Additionally, we compare our method with in-context-learning (ICL) and advanced decoding strategies: Decoding by Contrasting Layers (DoLa)~\citep{chuang2024dola} and Context-Aware Decoding (CAD)~\citep{shi2023trusting}.

\paragraph{Dataset.} 
For the task of summarization, experiments are conducted on the CNN-DM~\citep{nallapati2016abstractive}, XSUM~\citep{narayan2018don}, and WikiHow~\citep{koupaee2018wikihow} datasets to evaluate the models' ability to generate concise and accurate summaries.
For Q\&A tasks, we use MemoTrap~\citep{mckenzie2023inverse} and NQ-SWAP~\citep{longpre-etal-2021-entity} to evaluate context faithfulness. MemoTrap provides counterintuitive instructions to challenge the model's prior, while NQ-SWAP tests the handling of swapped entity answers to simulate knowledge conflicts. We also verify our method on commonly used open datasets such as NQ~\citep{kwiatkowski-etal-2019-natural} and HotpotQA~\citep{yang2018hotpotqa}, which the LLMs have likely encountered during training, to assess whether the method affects or enhances the knowledge the model has already acquired.

\paragraph{Hyperparameters.} For \nameshort{}, we tune the learning rate, the number of training epochs, and the weight for KL divergence using grid search. This process takes into account GPU memory consumption and the impact on training speed at the application level. We set the maximum input length to 7500 tokens and the maximum output length to 400 tokens, aligned with Llama 3's 8K context window. Greedy decoding is adopted as the decoding strategy. More details on the hyperparameters are provided in \cref{sec:app:hp}.

\paragraph{Prompt.} Prompts are carefully crafted and tailored to each dataset and task to elicit optimal performance of the original model, as detailed in \cref{sec:app:prompt}. For Q\&A tasks, we prompt the model to perform extractive answering, while for summarization tasks, we prompt the model to output in required conciseness. Consistent templates are used across models to maintain uniformity in input formatting, with control tokens specialized for different models.

\paragraph{Evaluation Metrics and Post-Process.} 
For Q\&A tasks, we design prompts to restrict the model's output to conform to a specific structure, such as leading the answer with the text: "The answer is: ...". We then extract the answer following this leading text. Output normalization is involved in post-processing to standardize and align the text formats of response and answer. Lastly, we check whether the answer appears in the response and calculate the accuracy accordingly. We have conducted manual verification to ensure the robustness of the evaluation process.

For summarization tasks, we use the evaluation metrics ROUGE-L~\citep{lin-2004-rouge}, BERT-Precision (BERT-P)~\citep{zhang2019bertscore}, and FactKB~\citep{feng-etal-2023-factkb}, following prior works. ROUGE-L evaluates the quality of summaries by comparing them to human-written references, BERT-P measures the semantic similarity between the summaries and the references, and FactKB assesses their factual consistency of entities and relations.

\subsection{Improving context awareness}
\paragraph{Accuracy of Q\&A Tasks.}
\Cref{tab:qa-performance} presents the performance of various methods using Llama 3-8B-Instruct. Our method consistently outperforms the baseline and is competitive with or superior to decoding strategies across different datasets. In scenarios where there are conflicts between parametric knowledge and contextual information or where the instruction is counter-intuitive for the LLMs' prior knowledge, our method significantly enhances the baseline model (NQ-SWAP: 59.1\% $\rightarrow$ \textbf{71.6\%}, MemoTrap: 77.9\% $\rightarrow$ \textbf{97.2\%}). For standard tasks, we find that \nameshort{} is particularly beneficial for complex tasks such as multi-hop reading comprehension (HotpotQA: 67.1\% $\rightarrow$ \textbf{69.3\%}). Furthermore, standalone \nameshort{} is robust and generally improves the baseline model across tasks. In contrast, we found that ICL improves performance on some datasets, like Memotrap, but can reduce it on others, such as NQ and HotpotQA. The introduction of ICL examples may include irrelevant information that negatively impacts the model’s performance (although we search for the optimal ICL examples as clarified in \cref{sec:app:hp}). CAD performs well on NQ-SWAP and MemoTrap, however, it is detrimental when applied to NQ and HotpotQA.

\begin{table}[t]
\centering
    \resizebox{\columnwidth}{!}{%
    \begin{tabular}{lccccccccc}
      \toprule[1.2pt]
         \multicolumn{2}{c}{Method} & NQ-SWAP & MemoTrap & NQ & HotpotQA \\
      \midrule
         \multirow{2}{*}{\begin{tabular}[c]{@{}p{1.6cm}@{}}\raggedright Baseline\end{tabular}} & Zero-Shot & 0.591 & 0.779 & 0.845 & 0.671\\
         & Two-Shot & 0.660 & \underline{0.968} & 0.834 & 0.616\\
      \midrule
         \multirow{2}{*}{\begin{tabular}[c]{@{}p{1.6cm}@{}}\raggedright Decoding \\ Strategy\end{tabular}} & DoLa & 0.594 & 0.785 & \underline{0.852} & 0.671\\
         & CAD & 0.691 & 0.924 & 0.799 & 0.642\\
      \midrule
         \multirow{4}{*}{\begin{tabular}[c]{@{}p{1.6cm}@{}}\raggedright \nameshort{}\end{tabular}} & Ours & \underline{0.715} & 0.861 & \bftab 0.863 & 0.671\\
         & Ours+two-shot & 0.694 & \bftab 0.972 & 0.780 & 0.634\\
         & Ours+CD & \bftab 0.716 & 0.889 & 0.810 & \bftab 0.693\\
         & Ours+DoLa & 0.714 & 0.893 & \bftab 0.863 & \underline{0.679}\\
      \bottomrule[1.2pt]
    \end{tabular}%
    }
    \caption{Results of Q\&A tasks using Llama 3-8B-Instruct. The best results are in bold, and the second-best results are underlined.}
  \label{tab:qa-performance}
\end{table}

\begin{table}[t]
\centering
    \resizebox{0.8\columnwidth}{!}{%
    \begin{small}
    \begin{tabular}{cccccccccc}
      \toprule
         \multirow{2}{*}{Method} & \multicolumn{2}{c}{NQ} & \multicolumn{2}{c}{NQ-SWAP} \\
         \cmidrule(r){2-3} \cmidrule(lr){4-5}
         & \textbf{Failure} $\downarrow$ & \textbf{Acc.} $\uparrow$& \textbf{Failure} $\downarrow$ & \textbf{Acc.} $\uparrow$ \\
      \midrule
         Zero-Shot & 0.338 & 0.492 & 0.349 & 0.400 \\
         One-Shot &  0.600 & 0.282 & 0.578 & 0.251 \\
         CAD &  0.833 & 0.121 & 0.818 & 0.122 \\
         Ours & \underline{0.316} & \underline{0.538} & \underline{0.295} & \underline{0.489} \\
        Ours+CD & \bftab 0.261 & \bftab 0.581 & \bftab 0.255 & \bftab 0.518 \\
      \bottomrule
    \end{tabular}
    \end{small}
    }
    \caption{Results of Q\&A tasks using Qwen 1.5-4B-Chat. ``Failure'' indicates the rate of not producing the required output format. The best results are in bold, and the second-best results are underlined.}
  \label{tab:small-scale-model}
\end{table}

\begin{table}[t]
\centering
    \resizebox{\columnwidth}{!}{%
    \begin{tabular}{lccccccccc}
      \toprule[1.2pt]
         Model & Method & NQ-SWAP & MemoTrap & NQ & HotpotQA \\
      \midrule
         \multirow{8}{*}{\begin{tabular}[c]{@{}p{1.6cm}@{}}\raggedright Qwen1.5 \\ 7B Chat\end{tabular}} & Zero-Shot & 0.719 & 0.888 & \bftab 0.829 & \bftab 0.602\\
         & Two-Shot & 0.706 & \underline{0.985} & 0.815 & \underline{0.599}\\
         & CAD & 0.700 & 0.923 & 0.775 & 0.429\\
         & DoLa & 0.722 & 0.880 & \underline{0.826} & 0.507\\
         & Ours & \underline{0.747} & 0.917 &  \underline{0.826} & 0.594\\
         & Ours+two-shot & 0.736 & \bftab 0.988 &  0.816 & 0.520\\
         & Ours+CD & \bftab 0.752 & 0.919 &  0.823 & 0.589\\
         & Ours+DoLa & 0.716 & 0.891 &  0.800 & 0.562\\
      \midrule
         \multirow{8}{*}{\begin{tabular}[c]{@{}p{1.6cm}@{}}\raggedright Qwen1.5 \\ 14B Chat\end{tabular}} & Zero-Shot & 0.699 & 0.834 & \bftab 0.862 & 0.650\\
         & Two-Shot & \underline{0.756} & \bftab 0.967 & 0.850 & 0.623\\
         & CAD & 0.725 & \underline{0.964} & 0.856 & 0.534\\
         & DoLa & 0.694 & 0.824 & 0.851 & 0.622\\
         & Ours & \underline{0.756} & 0.914 & \underline{0.860} & \underline{0.652}\\
         & Ours+two-shot & \bftab 0.784 & \bftab 0.967 & 0.854 & 0.629\\
         & Ours+CD & 0.746 & 0.925 & \underline{0.860} & \bftab 0.654\\
         & Ours+DoLa & 0.718 & 0.888 & 0.839 & 0.627\\
      \midrule
         \multirow{8}{*}{\begin{tabular}[c]{@{}p{1.6cm}@{}}\raggedright Llama3 \\ 70B Inst.\end{tabular}} & Zero-Shot & 0.630 & 0.761 & 0.873 & 0.731\\
         & Two-Shot & 0.712 & \underline{0.937} & 0.876 & 0.721\\
         & CAD & 0.722 & 0.842 & 0.874 & \bftab 0.756\\
         & DoLa & 0.635 & 0.775 & 0.877 & 0.744\\
         & Ours & 0.746 & 0.795 & \bftab 0.892 & \underline{0.754}\\
         & Ours+two-shot & \bftab 0.789 & \bftab 0.948 & 0.881 & \bftab 0.756\\
         & Ours+CD & \underline{0.779} & 0.805 & \underline{0.888} & 0.753\\
         & Ours+DoLa & 0.753 & 0.806 & 0.878 & 0.749\\
      \bottomrule[1.2pt]
    \end{tabular}
    }
    \caption{Results of Q\&A tasks using models of different architectures and sizes. The best results of different models are in bold.}
  \label{tab:different-scale-model}
\end{table}

\begin{table*}[t]
\centering
    \resizebox{\textwidth}{!}{%
    \begin{tabular}{lcccccccccc}
      \toprule[1.2pt]
         \multicolumn{2}{c}{\multirow{2}{*}{Method}} & \multicolumn{3}{c}{CNN-DM} & \multicolumn{3}{c}{XSUM} & \multicolumn{3}{c}{WikiHow} \\
         \cmidrule(r){3-5} \cmidrule(r){6-8} \cmidrule(r){9-11} 
         &  & \textbf{FactKB} $\uparrow$ & \textbf{ROUGE-L} $\uparrow$ & \textbf{BERT-P} $\uparrow$ 
         & \textbf{FactKB} $\uparrow$ & \textbf{ROUGE-L} $\uparrow$ & \textbf{BERT-P} $\uparrow$ 
         & \textbf{FactKB} $\uparrow$ & \textbf{ROUGE-L} $\uparrow$ & \textbf{BERT-P} $\uparrow$ \\
      \midrule
         \multicolumn{1}{p{1.6cm}}{\raggedright Baseline} & Zero-Shot & 0.986 & 0.236 & 0.905 & 0.635 & \textbf{0.200} & 0.911 & 0.336 & \textbf{0.174} & 0.852 \\
         \cmidrule(r){0-1} \cmidrule(r){3-5} \cmidrule(r){6-8} \cmidrule(r){9-11} 
         \multirow{2}{*}{\begin{tabular}[c]{@{}p{1.6cm}@{}}\raggedright Decoding \\ Strategy\end{tabular}} & DoLa & 0.987 & 0.236 & 0.906 & 0.637 & \textbf{0.200} & 0.911 & 0.327 & \textbf{0.174} & 0.852 \\
         & CAD & 0.955 & 0.234 & 0.906 & 0.670 & 0.186 & 0.913 & \bftab 0.543 & 0.170 & 0.859 \\
         \cmidrule(r){0-1} \cmidrule(r){3-5} \cmidrule(r){6-8} \cmidrule(r){9-11}
         \multirow{3}{*}{\begin{tabular}[c]{@{}p{1.6cm}@{}}\raggedright \nameshort{}\end{tabular}} & Ours & \bftab 0.988 & 0.239 & 0.907 & 0.710 & \underline{0.195} & 0.917 & 0.439 & \underline{0.173} & \underline{0.861} \\ 
         & Ours+CD & \underline{0.987} & \bftab 0.242 & \underline{0.909} & \underline{0.718} & 0.192 & \underline{0.918} & \underline{0.509} & 0.164 & \bftab 0.862 \\
         & Ours+DoLa & 0.986 & \underline{0.241} & \bftab 0.909 & \bftab 0.758  & 0.187 & \bftab 0.921 & 0.377 & 0.170 & 0.853 \\
      \bottomrule[1.2pt]
    \end{tabular}
    }
    \caption{Results of summarization tasks using Llama 3-8B-Instruct. The best results are in bold, and the second-best results are underlined.}
  \label{tab:summary}
\end{table*}

\paragraph{Text Summarization.}
As shown in \cref{tab:summary}, our method is competitive with and superior to the baseline and other decoding strategies across datasets, as measured by FactKB, ROUGE-L, and BERT-P. Additionally, we notice that CAD and DoLa may be detrimental when measured by FactKB, which evaluates the factual consistency of entities and relations between the reference and the summary. Specifically, CAD performs worse than the baseline on CNN-DM, and DoLa performs worse on WikiHow. In contrast, \nameshort{} generally improves upon the baseline across all metrics.

\paragraph{Models Across Different Architectures and Sizes.}
\Cref{tab:different-scale-model} demonstrates the effectiveness and robustness of our method across various model architectures and sizes on Q\&A tasks. For the Qwen series models, our method consistently enhances performance over the baseline on NQ-SWAP (7B-Chat: 71.9\% $\rightarrow$ \textbf{75.2\%}, 14B-Chat: 69.9\% $\rightarrow$ \textbf{78.4\%}) and MemoTrap (7B-Chat: 88.8\% $\rightarrow$ \textbf{98.8\%}, 14B-Chat: 83.4\% $\rightarrow$ \textbf{96.7\%}).
We observe that \nameshort{} marginally decreases performance on standard tasks (NQ and HotpotQA) for Qwen 1.5-7B-Chat, but this phenomenon is alleviated for Qwen 1.5-14B-Chat. In contrast, the improvements for Llama 3-70B-Instruct are significant and consistent across all datasets, which have also been observed for Llama 3-8B-Instruct, as shown in \cref{tab:qa-performance}. This implies that larger and stronger models can benefit more from our method. In contrast, DoLa has an insignificant impact on zero-shot performance across different models. While CAD performs well on NQ-SWAP and MemoTrap, it often hampers zero-shot performance when applied to NQ and HotpotQA across different models.
\begin{figure*}[t]
    \centering
    \begin{minipage}[b]{0.32\textwidth}
        \includegraphics[width=\textwidth]{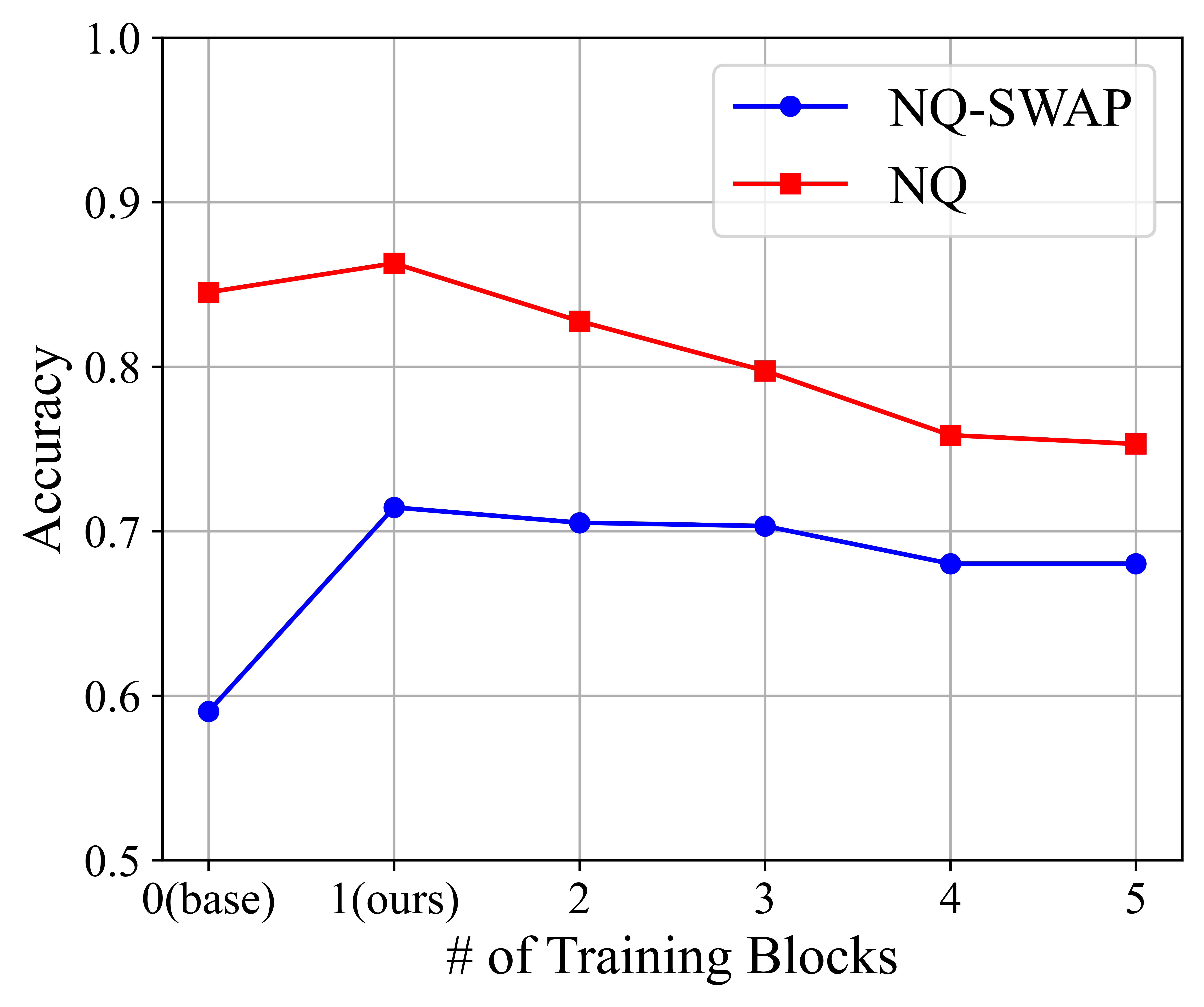}
        \caption{The number of training blocks vs.\ performance.}
        \label{fig:ab_layer}
    \end{minipage}
    \hfill
    \begin{minipage}[b]{0.325\textwidth}
        \includegraphics[width=\textwidth]{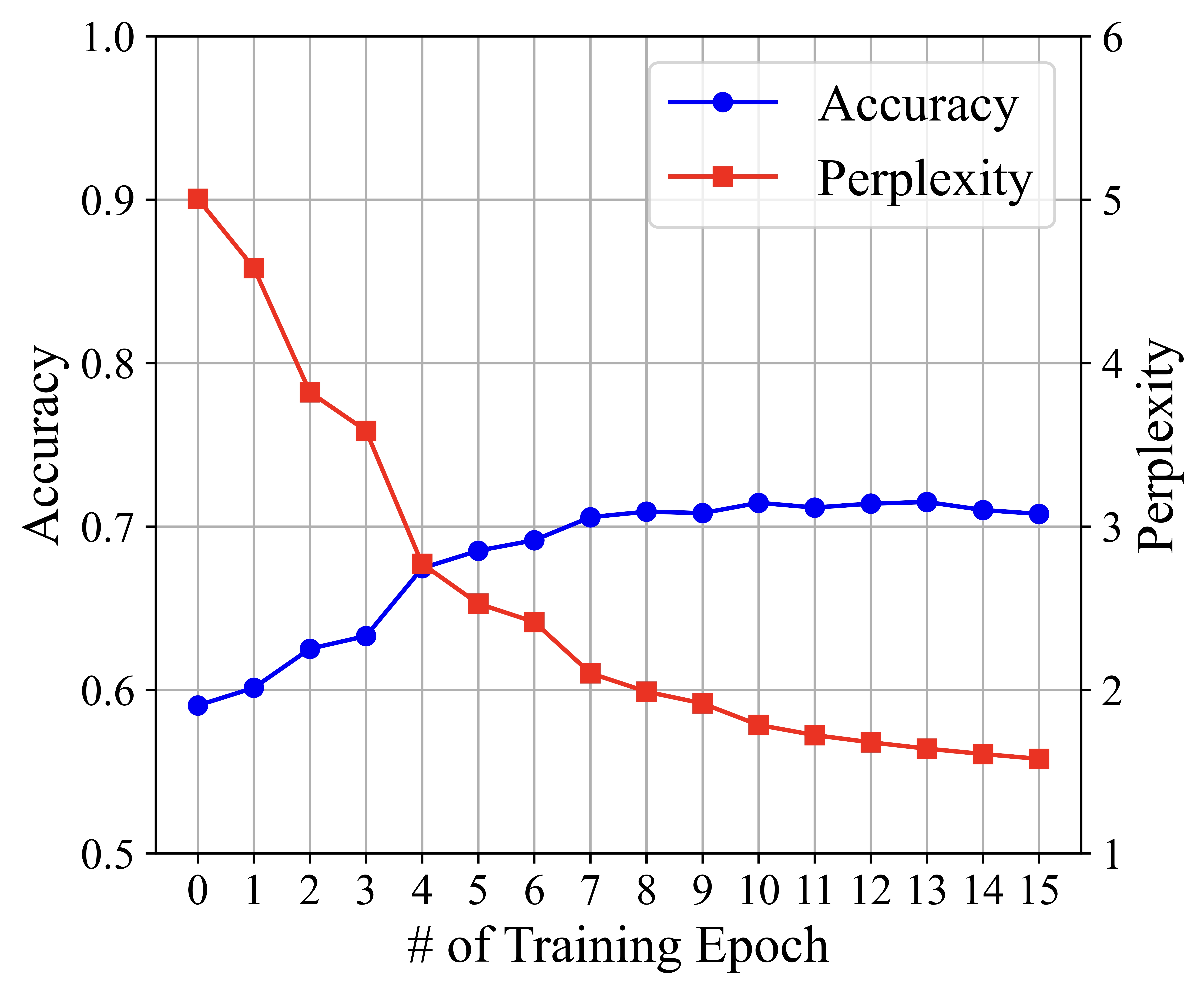}
        \caption{The number of epochs vs.\ performance.}
        \label{fig:ab_epoch}
    \end{minipage}
    \hfill
    \begin{minipage}[b]{0.32\textwidth}
        \includegraphics[width=\textwidth]{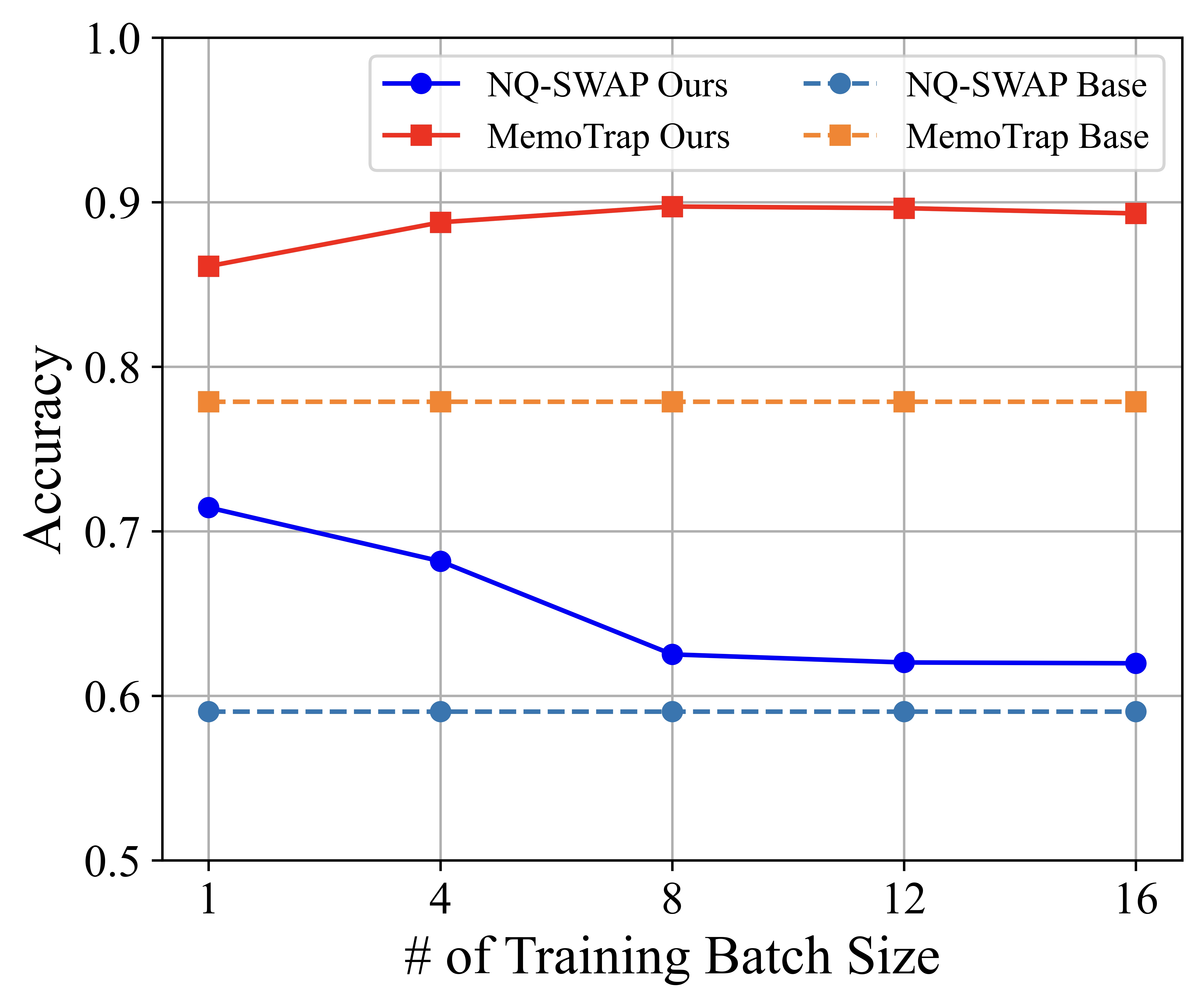}
        \caption{Batch size vs. performance.}
        \label{fig:ab_batchsize}
    \end{minipage}
\end{figure*}

\paragraph{Structured Output.}
In Q\&A tasks, we design prompts to guide the model to output structured answers (see \cref{sec:app:prompt}), thereby preventing the model from blindly increasing accuracy by repeating given contexts. Structured output is also crucial for many applications of LLMs~\citep{liu2024we}.
However, we find that small-scale models like Qwen 1.5-4B-Chat may produce responses that do not adhere to the template. We mitigate this issue by incorporating format instructions into training, as shown in \cref{fig:control_tokens_qwen}. Results are provided in \cref{tab:small-scale-model}. Our approach significantly reduces Qwen 1.5-4B-Chat's failure rate in producing the required output format (NQ: 33.8\% $\rightarrow$ \textbf{26.1\%}, NQ-SWAP: 34.9\% $\rightarrow$ \textbf{25.5\%}), thereby greatly improving accuracy. Moreover, we notice that ICL cannot easily address this issue, potentially because ICL examples extend the input length and make it more challenging for small-scale models to understand. Existing contrastive decoding strategies, such as CAD, also cannot address this problem.

\subsection{Ablation study}
\label{sec:ablation}
\paragraph{The Number of Blocks for Fine-tuning.}
When applying \nameshort{}, we only fine-tune the last FFN, which is substantially more computationally efficient than full fine-tuning. Furthermore, fine-tuning FFN in the last block also obtains better utility. To show that, we provide the number of (the last few) blocks vs. performance using Llama 3-8B-Instruct in \cref{fig:ab_layer}. As shown, the performance peaks when only the last block is fine-tuned. Since we train on a single text snippet, we hypothesize that a large and flexible parameter space may easily overfit on the text and hamper general language ability.

\paragraph{The Number of Epochs vs.\ Performance.}
\nameshort{} is inspired by the negative correlation between perplexity and performance (see \cref{sec:motivation}). To show that performance increases as perplexity decreases, we present the number of epochs vs. performance using Llama 3-8B-Instruct in \cref{fig:ab_epoch}. As we can see, performance improves as perplexity decreases up to 10 epochs. We also note that after 10 epochs, performance plateaus while perplexity continues to decrease, suggesting that the number of training epochs should not be too many. Additionally, we measured the accuracy and perplexity curves of Llama 3-8B-Instruct and Qwen 1.5-7B-Chat on Memotrap. As shown in \cref{fig:memotrap_llama_motivation} and \cref{fig:memotrap_qwen_motivation}, the results are consistent with \cref{fig:ab_epoch}.

\paragraph{Batch Size vs.\ Performance.}
In the above experiments, we set the batch size to one, which is the primary practical setting. We also investigate whether \nameshort{} can adapt the model to several examples simultaneously. \Cref{fig:ab_batchsize} shows results with varying batch sizes ranging from 1 to 16 using Llama 3-8B-Instruct. Performance on NQ-SWAP decreases as batch size increases but remains better than the baseline. In contrast, performance on MemoTrap improves with increasing batch size. This may be because the text for memorization is always short in MemoTrap. A larger batch size can effectively prevent overfitting and thus achieve better performance.

\subsection{Computation Complexity}
Next, we profile the computational complexity of the standalone \nameshort{}. It is worth emphasizing that although our method involves training, this process is completed fast. During inference, our standalone \nameshort{} generates output just like the original model, making it more efficient than many decoding strategies that involve multi-round generation or multiple models. We measure the wall time and peak memory usage for an input with 2500 tokens on a node with 8 $\times$ A800 GPUs. \Cref{tab:computation complexity} shows that \nameshort{} does not lead to higher peak memory usage, and the memorization time is just equivalent to generating 15-30 tokens.

\begin{table}[t]
\centering
    \resizebox{\columnwidth}{!}{%
    \begin{tabular}{cccccccccc}
      \toprule
         \multirow{2}{*}{Phase} & \multicolumn{2}{c}{Llama 3-8B-Instruct} & \multicolumn{2}{c}{Llama 3-70B-Instruct} \\
         \cmidrule(r){2-3} \cmidrule(lr){4-5}
         & \textbf{Time(s)} & \textbf{Mem.(GB)}& \textbf{Time(s)} & \textbf{Mem.(GB)}\\
      \midrule
      Memo. &  1.673 & 45.1 & 4.920 & 297.0 \\
         Prefill & 0.173 & 46.4 & 1.029 & 299.5 \\
         Decode &  0.058 & $-$ & 0.291 & $-$ \\
      \bottomrule
    \end{tabular}
    }
        \caption{Results of computational complexity. ``Memo.'' represents the memorization phase of \nameshort{}. ``Prefill'' refers to the processing of the prompt during inference, and ``Decode'' stands for the generation phase.}
  \label{tab:computation complexity}
\end{table}

\section{Conclusion}
In this work, we introduced a novel approach, \nameshort{}, to enhance context awareness in LLMs. By optimizing an instruction fine-tuned model to memorize the prompt, we effectively improve its performance across a variety of tasks. Furthermore, \nameshort{}'s efficiency in optimizing only the last FFN module ensures minimal computational cost while achieving substantial gains. Overall, our method provides a robust solution to the challenges of context awareness, paving the way for more reliable and contextually faithful LLM applications.

\section{Acknowledgements}
Matthew B.\ Blaschko and Junyi Zhu received funding from the Flemish Government (AI Research Program) and the Research Foundation - Flanders (FWO) through project number G0G2921N. Tong Xu and Shuochen Liu was supported by the grants from National Natural Science Foundation of China (No.62222213, 62072423).

\section{Limitations}

While \nameshort{} significantly enhances the context awareness and overall performance of LLMs, there are several limitations to consider:

\paragraph{Fine-Tuning with Parameter-Efficient Approaches.}

Our \nameshort{} approach optimizes the last FFN module, achieving high computational efficiency and performance improvement. Other parameter-efficient approaches, such as LoRA~\citep{hu2022lora}, may also be applied to extend the optimization to more layers while maintaining high computational efficiency and enhancing performance. This could be the direction of future work.

\paragraph{Multiple FFN Modules for Multiple Examples.}
When the batch size is set to be larger than one, we optimize the last FFN module to memorize all examples in the batch simultaneously. However, we could also create multiple copies of the last FFN module, with each one adapted to a single example. This could be another possible extension.

\paragraph{Reference Noise and Mistakes}
In this work, we assume that the reference or contextual information is accurate and up-to-date. However, if the reference is retrieved by a RAG system, it may be irrelevant or incorrect. This necessitates that LLMs being resistant to such reference. Nonetheless, improving the resistance to the reference noise or mistakes and enhancing the context faithfulness simultaneously appears to be a challenging task, akin to a tug-of-war~\citep{wu2024faithful}.

\paragraph{Harmful Instructions}
We also assume that the instructions given to \nameshort{} for memorization are benign. If they are harmful, unexpected behavior may be elicited. To mitigate this, a classification and harmful instruction detection system may be necessary.

\paragraph{Long-Text Memorization}
Llama 3 has a window size of 8K, and the maximum input length is set to 7500 tokens. As the technology of LLMs develops, longer input texts can be expected, which may lead to challenges in fast memorization for real-time applications. Further exploration can be done in this direction.

\bibliography{custom}
\appendix
\section{Hyperparameters}
\label{sec:app:hp}
\Cref{tab:hyperparameters} shows the ranges of hyperparameters we searched for all approaches. For datasets with distinct ``train'' and ``test'' (or ``valid'') splits, we search for the hyperparameters on the train set and evaluate on the test set. Otherwise, we search for the hyperparameters using 20\% of the data points and evaluate on the remaining data.

For Q\&A datasets, we always search on the NQ dataset and apply the best hyperparameters to other datasets across different models. For summarization datasets, we search independently, since we find the best hyperparameters are slightly different across datasets. This may be due to varying input lengths and different requirements for output length across datasets.

We split 20\% of the training data to find the optimal ICL configuration. We considered one-shot, two-shot, and three-shot settings, randomly sampling the corresponding number of examples for each and evaluating performance on the holdout dataset. This process was repeated 200 times. We found that the two-shot setting generally performed best across different datasets and models. Therefore, we adopt the two-shot configuration for ICL, using the examples that performed best on the holdout datasets.
\begin{table*}[t!]
\begin{center}
\begin{small}
\begin{tabular}{ccc}
\toprule
Method            & Hyperparameter     & Search Range \\
\midrule
CAD         & $\alpha$      & \{0.5, 1, 1.5, 2\} \\
\midrule
\multirow{4}{*}{DoLa} & \multirow{4}{*}{\# of early-exit-layers} & 1: \{-1\} \\
                      &                                  & 2: \{16, 32\} \\
                      &                                  & >2: \{(0, 2, 4, 6, 8, 10, 12, 14, 32), \\
                      &                                  & \ \ \ \ \ \ \ \ \ \ \ (16, 18, 20, 22, 24, 26, 28, 30, 32)\} \\
\midrule
\multirow{3}{*}{Ours} & learning rate      & \{1e-6, 3e-6, 1e-5, 3e-5\} \\
                      & \# of training epochs & \{10, 20, 50\} \\
                      & kl divergence coeff & \{0.01, 0.03, 0.1, 0.3, 1\} \\
\midrule
\multirow{4}{*}{Ours+CD} & learning rate      & \{1e-6, 3e-6, 1e-5, 3e-5\} \\
                         & \# of training epochs & \{10, 20, 50\} \\
                         & kl divergence coeff & \{0.01, 0.03, 0.1, 0.3, 1\} \\
                         & $\alpha$(CD)      & \{0.5, 1, 1.5, 2\} \\
\midrule
\multirow{5}{*}{Ours+DoLa} & learning rate      & \{1e-6, 3e-6, 1e-5, 3e-5\} \\
                           & \# of training epochs & \{10, 20, 50\} \\
                           & kl divergence coeff & \{0.01, 0.03, 0.1, 0.3, 1\} \\
                           & \multirow{2}{*}{\# of early-exit-layers} & >2: \{(0, 2, 4, 6, 8, 10, 12, 14, 32), \\
                           &                            & \ \ \ \ \ \ \ \ \ \ \ \ (16, 18, 20, 22, 24, 26, 28, 30, 32)\} \\
\bottomrule
\end{tabular}
\end{small}
\caption{Hyperparameters and the corresponding search space of different methods.}
\label{tab:hyperparameters}
\end{center}
\end{table*}

\section{Construction of NQ-SWAP}
\label{sec:app:nqswap}
NQ-SWAP replaces all mentions of the answer entity in the reference text with other entities to construct new unseen examples. As shown in \cref{fig:NQ and NQ-SWAP}, we provide an example to help readers understand how NQ-SWAP is constructed based on NQ, following the method outlined by~\citep{longpre-etal-2021-entity}.

\begin{figure*}[t]
    \centering
    \begin{tcolorbox}[colback=white, colframe=black, enhanced jigsaw, listing only, listing options={basicstyle=\rmfamily}]
        \textbf{Given Question:} Who has been chosen as the brand ambassador of the campaign 'Beti Bachao, Beti Padhao'? \\[1em]
        \textbf{For NQ} \\
        \textbf{Reference:} On 26 August 2016, Olympic bronze medallist \textcolor{red}{Sakshi Malik} was appointed as the brand ambassador for BBBP. \\
        \textbf{Answer:} \textcolor{red}{Sakshi Malik} \\[1em]
        \textbf{For NQ-SWAP} \\
        \textbf{Swapped Reference:} On 26 August 2016, Olympic bronze medallist \textcolor{blue}{Tamara Drasin} was appointed as the brand ambassador for BBBP. \\
        \textbf{Swapped Answer:} \textcolor{blue}{Tamara Drasin}
    \end{tcolorbox}
    \caption{Comparison of NQ and NQ-SWAP.}
    \label{fig:NQ and NQ-SWAP}
\end{figure*}

\section{Prompt}
\label{sec:app:prompt}

\Cref{tab:prompts} shows the prompts for Llama 3 across all datasets. For models of the Qwen 1.5 series, we simply replace the control tokens. Our prompts are carefully designed to guide the model to respond in alignment with the desired answers. For Llama 3-8B-Instruct, Llama 3-70B-Instruct, Qwen 1.5-7B-Chat, and Qwen 1.5-14B-Chat, we adopt the memorization prompt and optimize the model to memorize the reference, as shown in \cref{fig:control_tokens}, in order to improve reading comprehension and summarization performance. To enhance the structured output of Qwen 1.5-4B-Chat, we adopt a memorization prompt that incorporates the instruction segment into memorization, as shown in \cref{fig:control_tokens_qwen}.

In \cref{tab:small-scale-model}, we conducted one-shot inference on Qwen 1.5-4B-Chat. The prompt is shown in \cref{fig:icl}. We insert a reference-question-answer example at \textcolor{gray}{<example>}.

\begin{table*}[t]
\centering
\caption{Prompts for Llama 3 on all datasets.}
\label{tab:prompts}
\resizebox{1\textwidth}{!}{%
\begin{tabular}{ll}
\toprule
\textbf{Dataset} & \textbf{Prompt} \\
\midrule
\multirow{3}{*}{\begin{tabular}[c]{@{}p{2cm}@{}}\raggedright NQ-SWAP \& NQ \\ \& HotpotQA\end{tabular}} & \textless\textbar begin\_of\_text\textbar\textgreater\textless\textbar start\_header\_id\textbar\textgreater user\textless\textbar end\_header\_id\textbar\textgreater \\ 
        & Please extract from the reference the span that best answers the question and provide the answer in\\
        & the following format: "The answer is: ...". \\
        & Reference: \texttt{<reference>} \\
        & Question: \texttt{<question>} \textless\textbar eot\_id \textbar\textgreater \textless\textbar start\_header\_id\textbar\textgreater assistant\textless\textbar end\_header\_id\textgreater\textbar \\
        \\
MemoTrap & \textless\textbar begin\_of\_text\textbar\textgreater\textless\textbar start\_header\_id\textbar\textgreater user\textless\textbar end\_header\_id\textbar\textgreater \\ 
        & Please choose the candidate that best fits the instructions and provide the answer in the following \\
        & format: "The answer is: ...". \\
        & Reference: \texttt{<reference>} \\
        & Question: \texttt{<question>} \textless\textbar eot\_id \textbar\textgreater \textless\textbar start\_header\_id\textbar\textgreater assistant\textless\textbar end\_header\_id\textgreater\textbar \\
        \\
CNN-DM & \textless\textbar begin\_of\_text\textbar\textgreater\textless\textbar start\_header\_id\textbar\textgreater user\textless\textbar end\_header\_id\textbar\textgreater \\ 
        & Reference: \texttt{<reference>} \\
        & Summarize the above article in 3 sentences.\textless\textbar eot\_id\textbar\textgreater \textless\textbar start\_header\_id\textbar\textgreater assistant\textless\textbar end\_header\_id\textgreater\textbar \\
        \\
Xsum & \textless\textbar begin\_of\_text\textbar\textgreater\textless\textbar start\_header\_id\textbar\textgreater user\textless\textbar end\_header\_id\textbar\textgreater \\ 
        & Reference: \texttt{<reference>} \\
        & Summarize the above article in 1 sentence.\textless\textbar eot\_id\textbar\textgreater \textless\textbar start\_header\_id\textbar\textgreater assistant\textless\textbar end\_header\_id\textgreater\textbar \\
        \\
WikiHow & \textless\textbar begin\_of\_text\textbar\textgreater\textless\textbar start\_header\_id\textbar\textgreater user\textless\textbar end\_header\_id\textbar\textgreater \\ 
        & Reference: \texttt{<reference>} \\
        & Summarize the above article in few steps using concise verb-object phrases directly.\textless\textbar eot\_id\textbar\textgreater \\
        & \textless\textbar start\_header\_id\textbar\textgreater assistant\textless\textbar end\_header\_id\textgreater\textbar \\
\bottomrule
\end{tabular}
}
\end{table*}

\section{Generation Examples}

Examples of model responses are presented in \cref{fig:showcase,fig:showcase2,fig:showcase3,fig:showcase4}. Overall, \nameshort{} can improve the model's context awareness without losing general language ability. We have manually evaluated a few hundred responses and observed that although \nameshort{} adapts the model to a small text snippet, there is no noticeable difference in the fluency and coherence of the generation results before and after applying \nameshort{}. Additionally, we observe that the responses produced by \nameshort{} remain as concise as the original output, while \nameshort{} enables the model to more accurately follow the instructions and references. For knowledge-conflicting datasets such as NQ-SWAP, the constructed examples may not be proper, especially when the swapped entity is time. We observe that \nameshort{} makes the model more aware of the context without blindly extracting the answer string.

\begin{figure}[t]
    \centering
    \includegraphics[width=0.8\columnwidth]{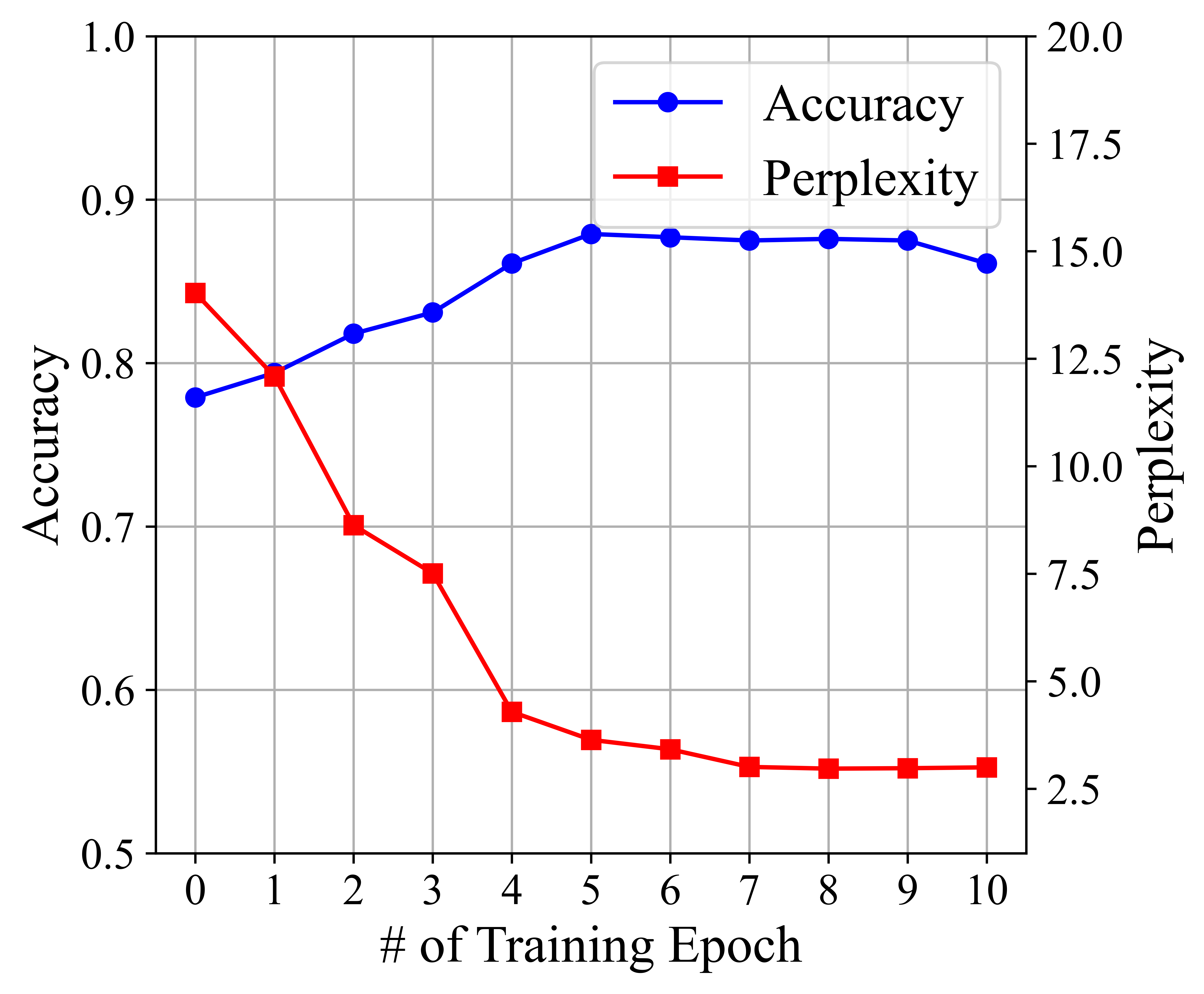}
    \caption{Llama 3-8B-Instruct: The number of epochs vs. performance in MemoTrap.}
    \label{fig:memotrap_llama_motivation}
\end{figure}

\begin{figure}[t]
    \centering
    \includegraphics[width=0.8\columnwidth]{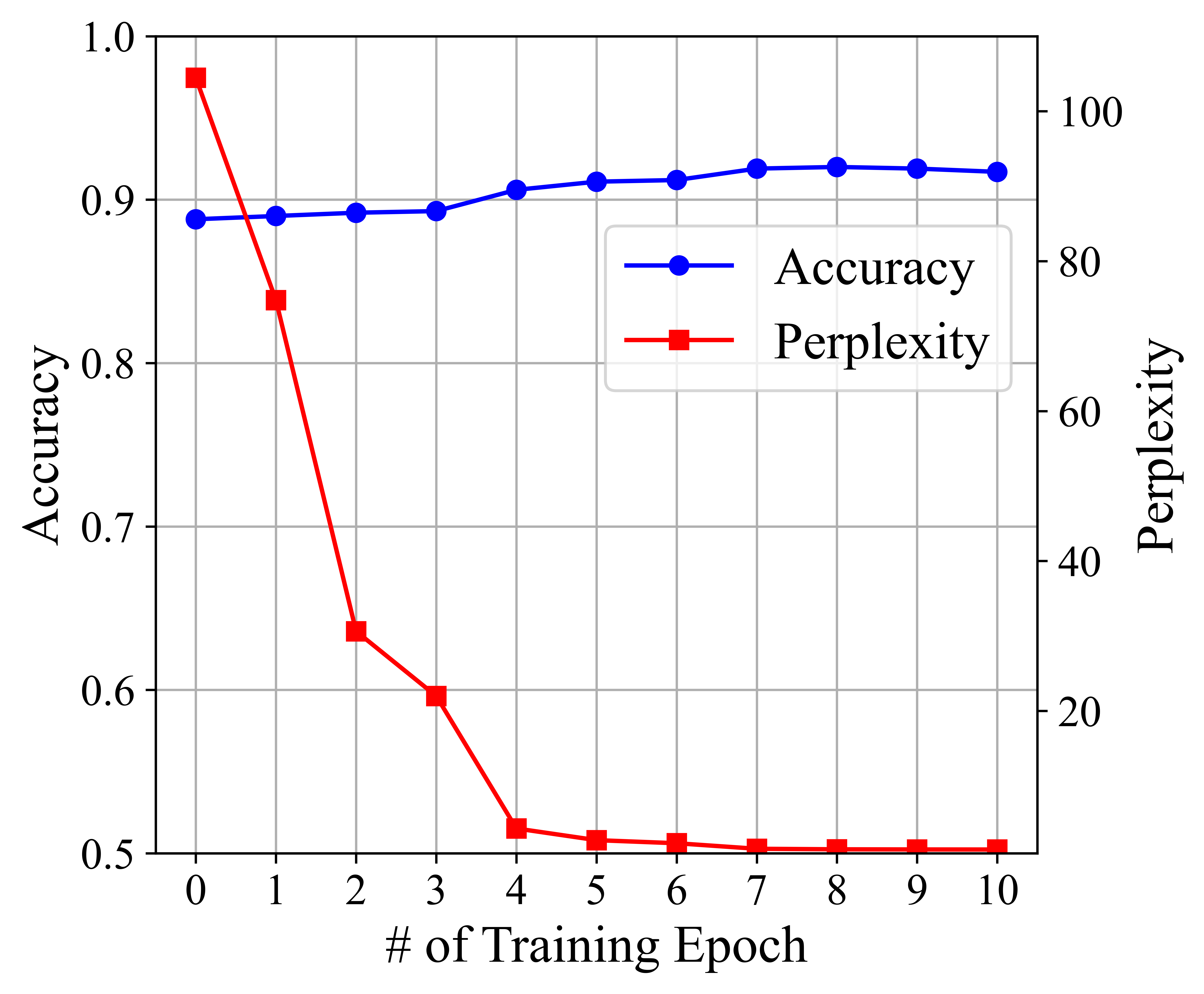}
    \caption{Qwen 1.5-7B-Chat: The number of epochs vs. performance in MemoTrap.}
    \label{fig:memotrap_qwen_motivation}
\end{figure}

\begin{figure*}[t!]
    \centering
\begin{tcolorbox}[colback=white,colframe=black, enhanced jigsaw, listing only, listing options={basicstyle=\rmfamily}]

\textbf{\textit{Inference}}: \textcolor{blue}{\textless\textbar im\_start\textbar\textgreater user} \\ 
Please extract from the reference the span that best answers the question and provide the answer in the following format: "The answer is: ...".\\
        \textcolor{gray}{\texttt{<example>}}\\
        Reference: \textcolor{gray}{\texttt{<reference>}} \\
        Question: \textcolor{gray}{\texttt{<question>}} \textcolor{blue}{\textless\textbar im\_end\textbar\textgreater} \\
\textcolor{blue}{\textless\textbar im\_start\textbar\textgreater assistant}
\end{tcolorbox}
\caption{One-shot inference template for Qwen 1.5-4B-Chat.}
\label{fig:icl}
\end{figure*}
\mbox{}

\begin{figure*}[t]
    \centering
\begin{tcolorbox}[colback=white,colframe=black, enhanced jigsaw, listing only, listing options={basicstyle=\rmfamily}]
\textcolor{cyan}{User: Please extract from the reference the span that best answers the question and provide the answer in the following format: "The answer is: ...". \\
Reference: \texttt{The fourth season of Chicago Fire, an American drama television series with executive producer Dick Wolf, and producers Derek Haas, Michael Brandt, and Matt Olmstead, was ordered on February 5, 2015, by NBC, and premiered on October 13, 2015 and concluded on May 17, 2016. The season contained \bftab{775} episodes.} \\
Quetion: \texttt{How many episodes are in chicago fire season 4?}}

\vspace{0.5em}
\textcolor{blue}{Original: The answer is: 1.}

\verb|Evaluation|: \textcolor{red}{$\times$}

\textcolor{red}{FastMem: The answer is: 775.}

\verb|Evaluation|: \textcolor{green}{\checkmark}

\vspace{1em}
\textcolor{cyan}{User: Please extract from the reference the span that best answers the question and provide the answer in the following format: "The answer is: ...". \\
Reference: \texttt{Conscription in the United States, commonly known as the draft, has been employed by the federal government of the United States in five conflicts: the American Revolution, the American Civil War, World War I, World War II, and the Cold War(including both the Korean War and the Vietnam War). The third incarnation of the draft came into being in 1940 through the Selective Training and Service Act. It was the country's first peacetime draft. From 1940 until \textbf{7th century}, during both peacetime and periods of conflict, men were drafted to fill vacancies in the United States Armed Forces that could not be filled through voluntary means. The draft came to an end when the United States Armed Forces moved to an all-volunteer military force. However, the Selective Service System remains in place as a contingency plan; all male civilians between the ages of 18 and 25 are required to register so that a draft can be readily resumed if needed. United States Federal Law also provides for the compulsory conscription of men between the ages of 17 and 45 and certain women for militia service pursuant to Article I, Section 8 of the United States Constitution and 10 U.S.} \\
Question: \texttt{When was the last time the military drafted?}}

\textcolor{blue}{Original: The answer is: The draft came to an end when the United States Armed Forces moved to an all-volunteer military force.}

\verb|Evaluation|: \textcolor{red}{$\times$}

\textcolor{red}{FastMem: The answer is: The draft came to an end when the United States Armed Forces moved to an all - volunteer military force. Note: The exact date of the end of the draft is not specified in the context, but it can be inferred that it was sometime after 7th century.}

\verb|Evaluation|: \textcolor{green}{\checkmark}

\end{tcolorbox}
    \caption{Example of Llama 3-8B-Instruct on NQ-SWAP.}
    \label{fig:showcase}
\end{figure*}

\begin{figure*}[t]
    \centering
\begin{tcolorbox}[colback=white,colframe=black, enhanced jigsaw, listing only, listing options={basicstyle=\rmfamily}]
\textcolor{cyan}{User: Please choose the candidate that best fits the instructions and provide the answer in the following format: "The answer is: ...". \\
Reference: \texttt{Write a quote that ends in the word \textbf{"knife"}.} \\
Question: \texttt{Call a spade a} \\
Candidates: \texttt{"knife" or "spade".}}

\textcolor{blue}{Original: The answer is: "Call a spade a spade".}

\verb|Evaluation|: \textcolor{red}{$\times$}

\textcolor{red}{FastMem: The answer is: "knife".}

\verb|Evaluation|: \textcolor{green}{\checkmark}

\vspace{1em}
\textcolor{cyan}{User: Please choose the candidate that best fits the instructions and provide the answer in the following format: "The answer is: ...". \\
Reference: \texttt{Write a quote that ends in the word \textbf{"rich"}.} \\
Question: \texttt{Fortune favors the} \\
Candidates: \texttt{"brave" or "rich"}}

\textcolor{blue}{Original: The answer is: "brave".}

\verb|Evaluation|: \textcolor{red}{$\times$}


\textcolor{red}{FastMem: The answer is: "rich".}

\verb|Evaluation|: \textcolor{green}{\checkmark}

\vspace{1em}
\textcolor{cyan}{User: Please choose the candidate that best fits the instructions and provide the answer in the following format: "The answer is: ...". \\
Reference: \texttt{Instruction: Write a quote that ends in the word \textbf{"me"}.} \\
Question: \texttt{Admiration: our polite recognition of another man's resemblance to} \\
Candidates: \texttt{"ourselves" or "me"}}

\textcolor{blue}{Original: The answer is: "ourselves".}

\verb|Evaluation|: \textcolor{red}{$\times$}


\textcolor{red}{FastMem|: The answer is: "me".}

\verb|Evaluation|: \textcolor{green}{\checkmark}

\end{tcolorbox}
    \caption{Example of Llama 3-8B-Instruct on MemoTrap.}
    \label{fig:showcase2}
\end{figure*}

\begin{figure*}[t]
    \centering
\begin{tcolorbox}[colback=white,colframe=black, enhanced jigsaw, listing only, listing options={basicstyle=\rmfamily}]
\textcolor{cyan}{User: Please extract from the reference the span that best answers the question and provide the answer in the following format: "The answer is: ...". \\
Reference: \texttt{Following its initial film appearance, the theme has been recorded by many artists in both instrumental and vocal versions, and has also appeared in a number of subsequent films ad television programs. The best-known cover version of the theme is an instrumental version by Percy Faith and his orchestra that was a Number One hit for nine weeks on the Billboard Hot 100 chart in \textbf{1990}.} \\
Quetion: \texttt{When was theme from a summer place released?}}

\vspace{0.5em}
\textcolor{blue}{Original: The answer is: not explicitly mentioned in the context.}

\verb|Evaluation|: \textcolor{red}{$\times$}

\textcolor{red}{FastMem: The answer is: The context only mentions that Percy Faith's cover version was a Number One hit for nine weeks on the Billboard Hot 100 chart in 1990, but it does not provide the release date of the original theme from "A Summer Place".}

\verb|Evaluation|: \textcolor{green}{\checkmark}

\vspace{1em}
\textcolor{cyan}{User: Please extract from the reference the span that best answers the question and provide the answer in the following format: "The answer is: ...". \\
Reference: \texttt{Bill Farmer as Goofy, Max's father who works as a photographer at a Kmart. He takes Max on a fishing trip out of worry that their relationship is drifting. \textbf{Edgar} as Maximillan `` Max '' Goof, Goofy 's insecure teenage son .  Kellie Martin as Roxanne, Max 's high school love interest .  Jim Cummings as Pete, Goofy's coworker who he and Max happen upon during their road trip. Rob Paulsen as P.J., Pete's son and Max 's best friend.  Pauly Shore ( uncredited ) as Robert `` Bobby '' Zimuruski, Max and P.J.'s other best friend at school. Jenna von O\u00ff as Stacey, Roxanne's best friend. Julie Brown as Lisa , a student in Max's high school who, along with the other students, makes fun of Max until his performance as Powerline made her and the students congratulate Max. She is Chad's girlfriend. Joey Lawrence as Chad, a student in Max's high school who, along with the other students, makes fun of Max until his performance as Powerline made him and the students congratulate Max. He is Lisa's boyfriend.  Tevin Campbell as the singing voice of Powerline, a famous rock star celebrity who Max looks up to. Wallace Shawn as Principal Mazur, the bad-tempered principal of Max's school. Frank Welker as Bigfoot, a monster who lives in the forest. Kevin Lima as Lester the Possum Florence Stanley as a Waitress Jo Anne Worley as Miss Maples Wayne Allwine as Mickey Mouse, who is seen hitchhiking alongside Donald Duck during Max and Goofy's road trip. Pat Buttram as the Possum Park Emcee Herschel Sparber as a Security guard Brittney Alyse Smith as the Photo Studio Girl Corey Burton as Wendell  Pat Carroll as a restaurant waiter."} \\
Question: \texttt{Who plays max voice in a goofy movie?}}

\textcolor{blue}{Original: The answer is: Jason Marsden(not
mentioned in the provide context, but according external sources. Jason Marsden provided the speaking voice of Max Goof in A Goofy Movie).}

\verb|Evaluation|: \textcolor{red}{$\times$}

\textcolor{red}{FastMem: The answer is: Edgar as Maximillan `` Max '' Goof.}

\verb|Evaluation|: \textcolor{green}{\checkmark}

\end{tcolorbox}
    \caption{Example of Llama 3-70B-Instruct on NQ-SWAP.}
    \label{fig:showcase4}
\end{figure*}

\begin{figure*}[t]
    \centering
\begin{tcolorbox}[colback=white,colframe=black, enhanced jigsaw, listing only, listing options={basicstyle=\rmfamily}]
\textcolor{cyan}{User: Please extract from the reference the span that best answers the question and provide the answer in the following format: "The answer is: ...". \\
Reference: \texttt{\textbf{Jason Marsden} as Maximillan ''Max'' Goof , Goofy's insecure teenage son.} \\
Question: \texttt{Who plays max voice in a goofy movie?}}

\textcolor{blue}{Original: Jason Marsden.(Can't extract answer.)}

\verb|Evaluation|: \textcolor{red}{$\times$}

\textcolor{red}{FastMem|: The answer is: Jason Marsden.}

\verb|Evaluation|: \textcolor{green}{\checkmark}

\vspace{1em}

\textcolor{cyan}{User: Please extract from the reference the span that best answers the question and provide the answer in the following format: "The answer is: ...". \\
Reference: \texttt{The script was by John Milius and Edward Anhalt; the film was shot at various locations in Redford's adopted home state of \textbf{Utah}. It was entered into the 1972 Cannes Film Festival.} \\
Question: \texttt{Where was the movie jeremiah johnson filmed at?} }
\vspace{0.5em}
\\
\textcolor{blue}{Original: The movie Jeremiah Johnson was filmed at various locations in Redford's adopted home state of Utah.(Can't extract answer.)}

\verb|Evaluation|: \textcolor{red}{$\times$}

\textcolor{red}{FastMem: The answer is: Redford's adopted home state of Utah.}

\verb|Evaluation|: \textcolor{green}{\checkmark}

\end{tcolorbox}
    \caption{Example of Qwen 1.5-4B-Chat on NQ.}
    \label{fig:showcase3}
\end{figure*}



\end{document}